\documentclass[runningheads]{llncs}
\usepackage{graphicx}
\usepackage{amsmath,amssymb} 
\usepackage{color}
\usepackage[width=122mm,left=12mm,paperwidth=146mm,height=193mm,top=12mm,paperheight=217mm]{geometry}

\usepackage{amsmath,amssymb}
\usepackage{multirow}
\usepackage{tabularx}

\begin{document}

	\pagestyle{headings}
	\mainmatter

	\title{Distinctive-attribute Extraction for Image Captioning}
	\authorrunning{B. Kim et al.}
	\author{Boeun Kim, Young Han Lee, Hyedong Jung and Choongsang Cho}
	\institute{Artificial Intelligence Research Center,\\ Korea Electronics Technology Institute, Korea }

\maketitle

\begin{abstract}
Image captioning, an open research issue, has been evolved with the progress of deep neural networks.
Convolutional neural networks (CNNs) and recurrent neural networks (RNNs) are employed to compute image features and generate natural language descriptions in the research. In previous works, a caption involving semantic description can be generated by applying additional information into the RNNs. 
In this approach, we propose a distinctive-attribute extraction (DaE) which explicitly encourages significant meanings to generate an accurate caption describing the overall meaning of the image with their unique situation. Specifically, the captions of training images are analyzed by term frequency-inverse document frequency (TF-IDF), and the analyzed semantic information is trained to extract distinctive-attributes for inferring captions. The proposed scheme is evaluated on a challenge data, and it improves an objective performance while describing images in more detail.
\keywords{Image captioning, Semantic information, Distinctive-attribute, and Term frequency-inverse document frequency (TF-IDF)}
\end{abstract}

\section{Introduction}

Automatically to describe or explain the overall situation of an image, an image captioning scheme is a very powerful and effective tool~\cite{vinyals2015show,donahue2015long,xu2015show}. The issue is an open research area in computer vision and machine learning~\cite{vinyals2015show,donahue2015long,xu2015show,gan2017semantic,dai2017contrastive,liu2017attention}.
In recent years, recurrent neural networks (RNNs) implemented by long short-term memory (LSTM) especially show good performances in sequence data processing and they are widely used as decoders to generate a natural language description from an image in many methods~\cite{xu2015show,gan2017semantic,dai2017contrastive,liu2017attention,hochreiter1997long}. 
High-performance approaches on convolutional neural networks (CNNs) have been proposed~\cite{simonyan2014very,he2016deep}, which are employed to represent the input image with a feature vector for the caption generation~\cite{xu2015show,gan2017semantic,dai2017contrastive}.

Additionally, an attention representation that reflects the human visual system has been applied to obtain salient features from an entire image~\cite{xu2015show}. The approach adopted in previous work provides different weights in an image effectively. 
High-level semantic concepts of the image are effective to describe a unique situation and a relation between objects in an image~\cite{gan2017semantic,karpathy2015deep}. Extracting specific semantic concepts encoded in an image, and applying them into RNN network has improved the performance significantly~\cite{gan2017semantic}. Detecting semantic attributes are a critical part because the high-level semantic information has a considerable effect on the performance. A recent work applied contrastive learning scheme into image captioning to generate distinctive descriptions of images~\cite{dai2017contrastive}.  

In this paper, we propose a Distinctive-attribute Extraction (DaE) which explicitly encourages semantically unique information to generate a caption that describes a significant meaning of an image. Specifically, it employs term frequency-inverse document frequency (TF-IDF) scheme~\cite{aizawa2003information} to evaluate a semantic weight of each word in training captions. The distinctive-attributes of images are predicted by a model trained with the semantic information, and then they are applied into RNNs to generate descriptions.

The main contributions of this paper are as follows: (i) We propose the semantics extraction method by using the TF-IDF caption analysis. (ii) We propose a scheme to compute distinctive-attribute by the model trained with semantic information.
(iii) We perform quantitative and qualitative evaluations,
demonstrating that the proposed method improves the performance of a base caption generation model by a substantial margin while describing images more distinctively.

This manuscript is organized as follows: In Section~\ref{sec:relatedWork},
the related schemes are explained. The proposed
scheme and its implementation are described in
Section~\ref{sec:proposed}, and the experimental results
are compared and analyzed in Section~\ref{sec:Experiment}. Finally,
in Section~\ref{sec:conclusion}, the algorithm is summarized, and a conclusion and discussions are presented.

\section{Related Work} 
\label{sec:relatedWork}
Combinations of CNNs and RNNs have been widely used for the image captioning networks~\cite{vinyals2015show,donahue2015long,xu2015show,gan2017semantic,fang2015captions,wu2016value}. An end-to-end neural network consisting of a vision CNN followed by a language generating RNN was proposed~\cite{vinyals2015show}. 
CNN was used as an image encoder, and an output of its last hidden layer is fed into the RNN decoder that generates sentences. 
Donahue \textit{et al.}~\cite{donahue2015long} proposed Long-term Recurrent Convolutional Networks(LRCN), which can be employed to visual time-series modeling such as generation of description. 
LRCN also used outputs of a CNN as LSTM inputs, which finally produced a description.

Recent approaches can be grouped into two paradigms.
Top-down includes attention-based mechanisms, and many of the bottom-up methods used semantic concepts. 
As approaches using the attention, Xu \textit{et al.}~\cite{xu2015show} introduced an attention-based captioning model, which can attend to salient parts of an image while generating captions. 
Liu \textit{et al.}~\cite{liu2017attention} tried to correct attention maps by human judged region maps. 
Different levels of correction were made dependent on an alignment between attention map and the ground truth region. 
Some other works extracted semantic information and applied them as additional inputs to the image captioning networks. 
Fang \textit{et al.}~\cite{fang2015captions} used Multiple Instance Learning (MIL) to train word detectors with words that commonly occur in captions, including nouns, verbs, and adjectives. 
The word detector outputs guided a language model to generate description to include the detected words. 
Wu \textit{et al.}~\cite{wu2016value} also clarified the effect of the high-level semantic information in visual to language problems such as the image captioning and the visual question answering. 
They predicted attributes by treating the problem as a multi-label classification. 
The CNN framework was used, and outputs from different proposal sub-regions are aggregated. 
Gan \textit{et al.}~\cite{gan2017semantic} proposed Semantic Concept Network (SCN) integrating semantic concept to a LSTM network. SCN factorized each weight matrix of the attribute integrated the LSTM model to reduce the number of parameters.
We employed SCN-LSTM as a language generator to verify the effectiveness of our method.

More recently, Dai \textit{et al.}~\cite{dai2017contrastive} studied the distinctive aspects of the image description that had been overlooked in previous studies. 
They said that distinctiveness is closely related to the quality of captions, 
The proposed method Contrastive Learning(CL) explicitly encouraged the distinctiveness of captions, while maintaining the overall quality of the generated captions. 
In addition to true image-caption pairs, this method used mismatched pairs which include captions describing other images for learning.

Term frequency-inverse document frequency(TF-IDF) is widely used in text mining, natural language processing, and information retrieval. 
TF indicates how often a word appears in the document. This measure employs a simple assumption that frequent terms are significant~\cite{aizawa2003information,luhn1957statistical}. 
A concept of IDF was first introduced as ``term specificity" by Jones~\cite{sparck1972statistical} in 1972. 
The intuition was a word which occurs in many documents is not a good discriminator and should be given small weight~\cite{sparck1972statistical,robertson2004understanding}.
Weighting schemes are often composed of both TF and IDF terms. 

\section{Distinctive-attribute Extraction}
\label{sec:proposed}

In this paper, we describe the semantic information processing and extraction method, which affects the quality of generated captions.
Inspired by the concept of Contrastive Learning (CL)~\cite{dai2017contrastive}, we propose a method to generate captions that can represent the unique situation of the image. However, different from CL that improved target method by increasing the training set, our method lies in the bottom-up approaches using semantic attributes. 
We assign more weights to the attributes that are more informative and distinctive to describe the image. 

\subsection{Overall Framework}
\label{subsec:overall_framework}

\begin{figure}
\centering
\includegraphics[width=12cm]{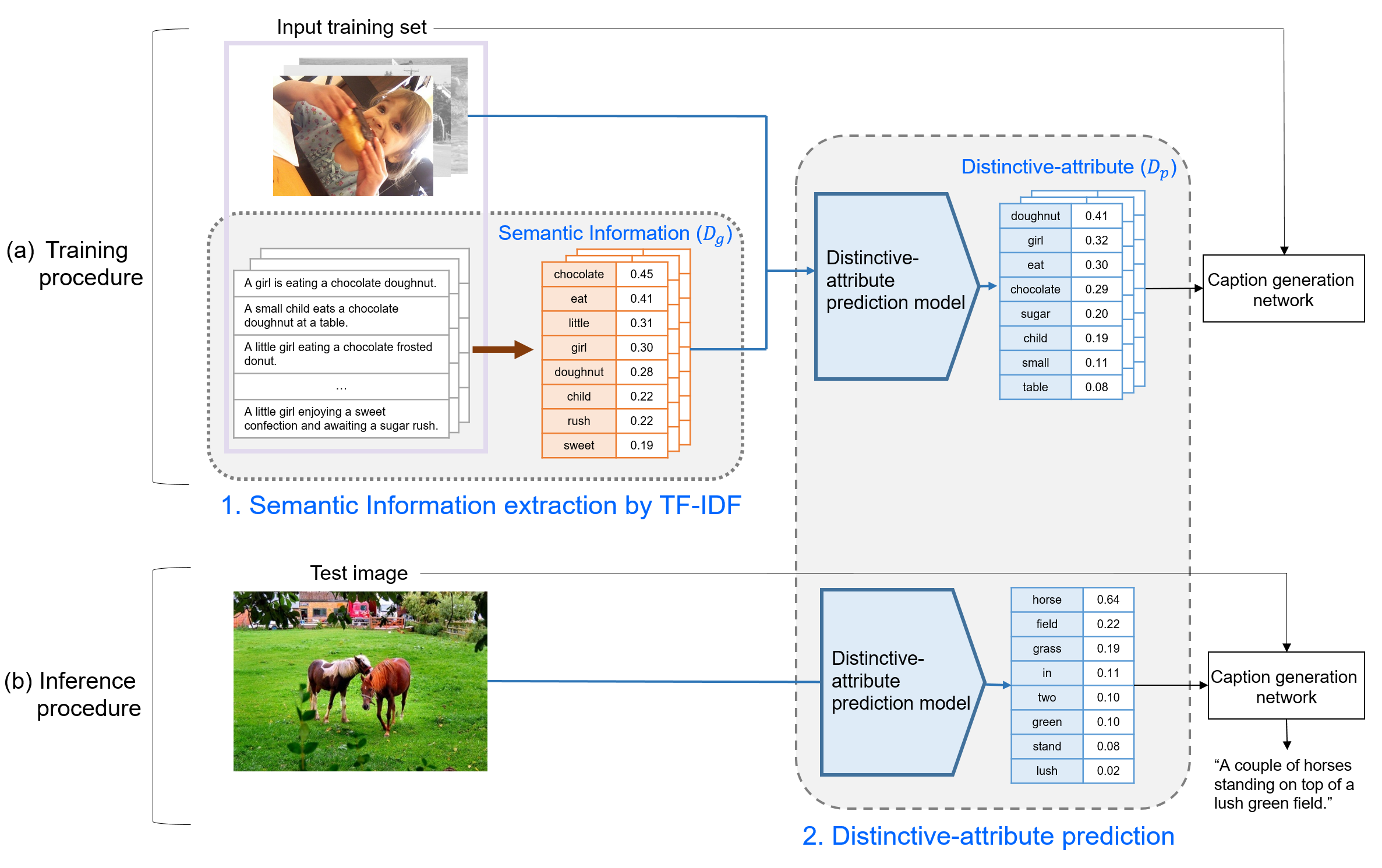}
\caption{An overview of the proposed framework. Given training captions, we extract semantically unique information. We employ a CNN-based model to predict distinctive-attribute. This attribute and the image feature are fed into the LSTM-based model and generate accurate captions.}
\label{fig:overall_framework}
\end{figure}

In this section, we explain overall process of our Distinctive-attribute Extraction(DaE) method.
As illustrated in Figure~\ref{fig:overall_framework}, there are two main steps, one is semantic information extraction, and the other is the distinctive-attribute prediction. 
We use TF-IDF scheme to extract meaningful information from reference captions.
In Section~\ref{subsec:extraction}, the method is discussed in detail and it contains a scheme to construct a vocabulary from the semantic information. After extracting the semantic information from training sets, we learn distinctive-attribute prediction model with image-information pairs. The model will be described in Section~\ref{subsec:model}. 
After getting distinctive-attribute from images, we apply these attributes to an caption generation network to verify their effect. 
We used SCN-LSTM~\cite{gan2017semantic} as a decoder which is a tag integrated network. Image features and distinctive-attributes predicted by the proposed model are served as inputs of the model. The SCN-LSTM unit with attribute integration and factorization~\cite{memisevic2007unsupervised} is represented as
\begin{align}
  {i}_{t} = \sigma({W}_{ia}\tilde{x}_{i,t-1} + {U}_{ia}\tilde{h}_{i,t-1} + z),
  \; \\
  {f}_{t} = \sigma({W}_{fa}\tilde{x}_{f,t-1} + {U}_{fa}\tilde{h}_{f,t-1} + z),
  \; \\
  {o}_{t} = \sigma({W}_{oa}\tilde{x}_{o,t-1} + {U}_{oa}\tilde{h}_{o,t-1} + z),
  \; \\
  \tilde{c}_{t} = \sigma({W}_{ca}\tilde{x}_{c,t-1} + {U}_{ca}\tilde{h}_{c,t-1} + z),
  \; \\
  {c}_{t} = {i}_{t}\odot\tilde{c}_{t} + {f}_{t}\odot{c}_{t-1},
  \; \\
  {h}_{t} = {o}_{t} \odot tanh({c}_{t}),
\end{align}
where $z = 1$ $(t=1)\cdot{C}_{v}$. $\odot$ denotes the element-wise multiply operator. For $\star = i,f,o,c$,
\begin{align}
  \tilde{x}_{\star, t-1} = {W}_{\star b}D_p \odot {W}_{\star c}{x}_{t-1},
  \; \\
  \tilde{h}_{\star, t-1} = {U}_{\star b}D_p \odot {U}_{\star c}{h}_{t-1},
\end{align}
where $D_p$ indicates distinctive-attribute predicted by the proposed model described in Section~\ref{subsec:model}. Similar to~\cite{gan2017semantic,wu2016value,menand2009show}, the objective function is composed of the conditional log-likelihood on the image feature and the attribute as 
\begin{align} \label{eq:log-likelyhood}
    p(X|I_n) =  \sum_{n=1}^{N} \log{p(X | f(I_{n}), D_p)}
\end{align}
where $I_{n}$, $f\left(  \cdot  \right)$, and $X$ indicates the n$th$ image, an image feature extraction function, and the caption, respectively. $N$ denotes the number of training images. The length$-T$ caption, $X$, is represented by a sequence of words; ${x}_{0}$, ${x}_{1}$, ${x}_{2}$, $\dots, {x}_{T}$. Modeling joint probability over the words with chain rule, log term is redefined as
\begin{align}
    \log{p(X | f(I), D_p)} = \sum_{t=1}^{T}\log{p({x}_{t} | {x}_{0}, \dots ,{x}_{t-1}, f(I), D_p)}.
\end{align}

\subsection{Semantic Information Extraction by TF-IDF}
\label{subsec:extraction}

Most of the previous methods constituted semantic information, that was a ground truth attribute, as a binary form~\cite{gan2017semantic,fang2015captions,wu2016value,you2016image}. 
They first determined vocabulary using K most common words in the training captions. 
The vocabulary included nouns, verbs, and adjectives. 
If the word in the vocabulary existed in reference captions, the corresponding element of an attribute vector became 1. 
Attribute predictors found probabilities that the words in the vocabulary are related to given image.

Different from previous methods, we weight semantic information according to their significance. 
There are a few words that can be used to describe the peculiar situation of an image. 
They allow one image to be distinguished from others. 
These informative and distinctive words are weighted more, and the weight scores are estimated from reference captions. 
We used the TF-IDF scheme which was widely used in text mining tasks for extracting the semantic importance of the word. 
Captions are gathered for each image, for example, five sentences are given in MS COCO image captioning datasets~\cite{lin2014microsoft,chen2015microsoft}, and they are treated as one document. 
The total number of documents must be the same as the number of images on a dataset.

\begin{figure}
\centering
\includegraphics[width=11cm]{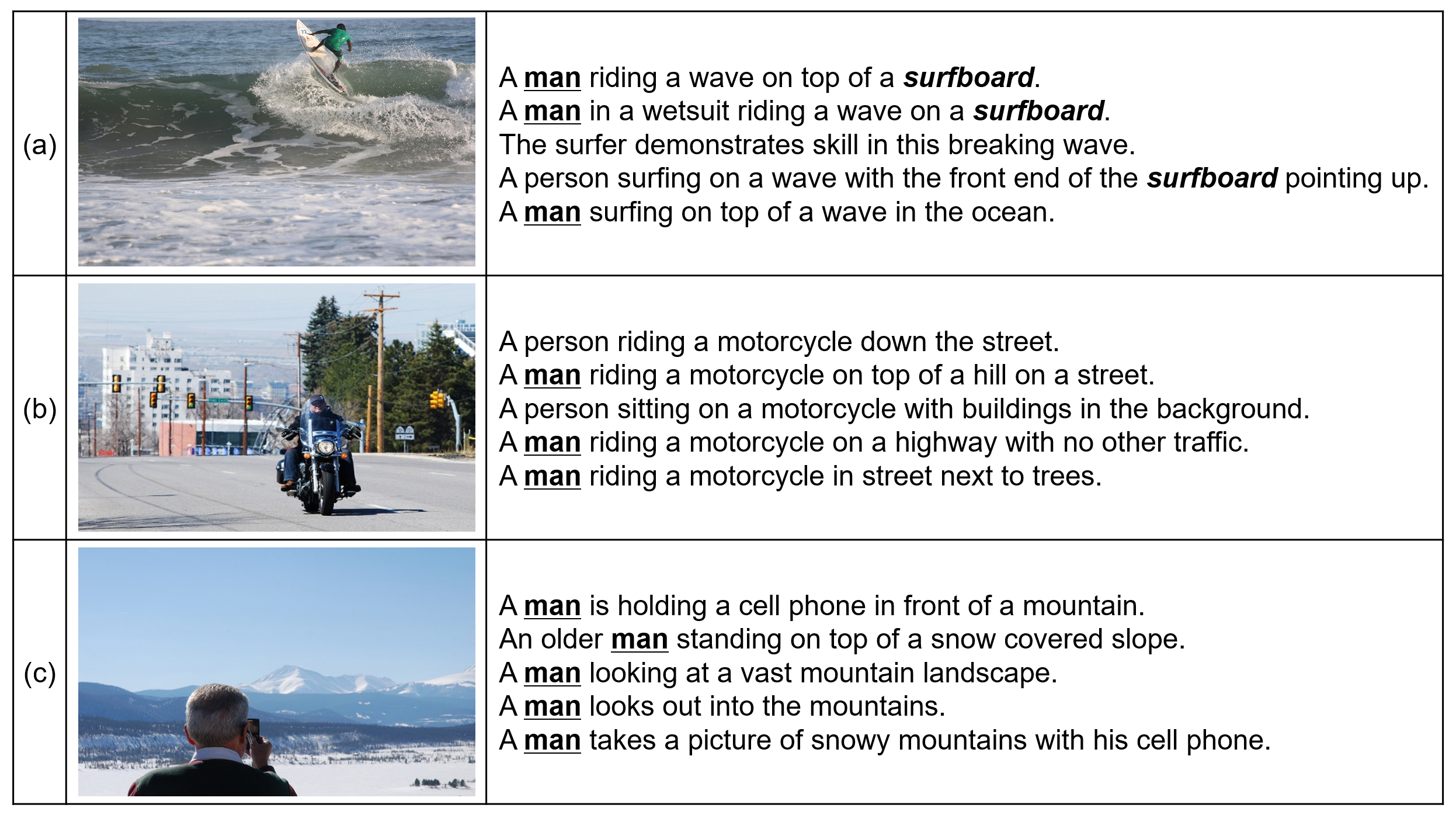}
\caption{Examples of images and their reference captions brought from MS COCO datasets~\cite{lin2014microsoft,chen2015microsoft}}
\label{fig:dataset_example}
\end{figure}

Figure~\ref{fig:dataset_example} represents samples of COCO image captioning, pairs of an image and captions. In \ref{fig:dataset_example}(a), there is a common word ``surfboard" in 3 out of 5 captions, which is a key-word that characterizes the image. Intuitively, this kind of words should get high scores. We apply TF to implement this concept and use average TF metric ${TF}_{av}$ which is expressed as 
\begin{align}
{TF}_{av}(w,d) = \frac{TF(w,d)}{{N}_{c}}
\end{align}
where $TF(w,d)$ denotes the number of times a word $w$ occurs in a document $d$. We divide $TF(w,d)$ by ${N}_{c}$ which is the number of captions for an image.

There is another common word ``man" in captions in Figure~\ref{fig:dataset_example}(a).
TF score of the word ``man" must be same as that of the word ``surfboard" because it appears 3 times. However, ``man" appears a lot in other images. 
Therefore, that is a less meaningful word for distinguishing one image from another. 
To reflect this, we apply inverse document frequency (IDF) term weighting. 
IDF metric for the word $w$ can be written as
\begin{align}
\label{eq:IDF}
  IDF(w) = \log{\frac{{N}_{d}+1}{DF(w)+1}}+1
\end{align}
where ${N}_{d}$ is the total number of documents, and $DF(w)$ is the number of documents that contain the word $w$. ``1" is added in denominator and numerator to prevent zero-divisions~\cite{pedregosa2011scikit}.
Then TF-IDF is derived by multiplying two metrics as
\begin{align}
TF-IDF(w,d) = {TF}_{av}(w,d) \times IDF(w).
\end{align}
We apply L2 normalization to TF-IDF vectors of each image for training performance. 
Consequently, the values are normalized into the range of 0 and 1. 
The semantic information vector which is the ground truth distinctive-attribute vector can be represented as
\begin{align}
{D}_{g,iw} = \frac{TF-IDF(w,d)}{ \lVert {TF-IDF}(w,d) \rVert_{2}}
\end{align}
where ${D}_{g,iw}$ indicates ground truth $D$ for image index $i$ and for word $w$ in vocabulary. $d$ denotes a document which is a set of reference captions for an image.

The next step is to construct vocabulary with the words in captions. 
It is essential to select the words that make up the vocabulary which ultimately affects captioning performance. 
The vocabulary should contain enough particular words to represent each image. 
At the same time, the semantic information should be trained well for prediction accuracy. 
In the perspective of vocabulary size, Gan~\cite{gan2017semantic} and Fang~\cite{fang2015captions} selected 1000 words and Wu~\cite{wu2016value} selected 256 words, respectively. 
They all selected vocabulary among nouns, verbs, and adjectives. 

We determine the words to be included in the vocabulary based on the IDF scores. We do not distinguish between verbs, nouns, adjectives, and other parts of speech. 
The larger the IDF value of a word is, the smaller the number of documents, i.e., the number of image data, which include the word. 
In this case, the word is said to be unique, but a model with this kind of inputs is challenging to be trained. 
We observe the performance of the semantic attribute prediction model and overall captioning model while changing the IDF value threshold. 

In addition, we compare the results with applying stemming before extracting TF-IDF. 
We assume that words with the same stem mostly mean same or relatively close concepts in a text. For example, ``looking" and ``looks" are mapped to the same word ``look" after stemming. Wu~\cite{wu2016value} did a similar concept, manually changing their vocabulary to be not plurality sensitive. We used Porter Stemmer algorithm~\cite{porter1980algorithm} which is implemented in Natural Language Toolkit (NLTK)~\cite{bird2004nltk}.

\subsection{Distinctive-attribute Prediction Model}
\label{subsec:model}

\begin{figure}
\centering
\includegraphics[width=11cm]{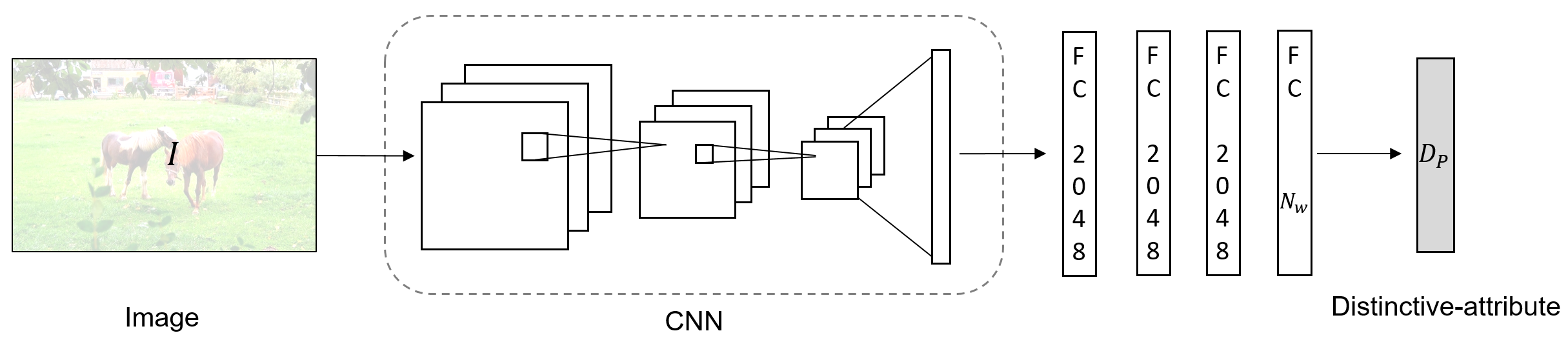}
\caption{A structure of distinctive-attribute prediction model. Convolutional layers are followed by four fully-connected layers}
\label{fig:prediction_model}
\end{figure}
For each image, distinctive-attribute vectors are inferred by a prediction model. 
Figure~\ref{fig:prediction_model} summarizes the distinctive-attribute prediction network. We use ResNet-152~\cite{he2016deep} architecture for CNN layers which have been widely used in vision tasks. 
The output of the 2048-way pool5 layer from ResNet-152~\cite{he2016deep} is fed into a stack of fully connected layers. 
This ResNet output is also reused in SCN-LSTM network as described in Section~\ref{subsec:overall_framework}. 
Training data for each image consist of input image $I$ and ground truth distinctive-attribute ${\textbf{D}}_{g,i} = [{D}_{g,i1}, {D}_{g,i2}, \dots , {D}_{g,i{N}_{w}}]$, where ${N}_{w}$ is the number of the words in vocabulary and $i$ is the index of the image. 
Our goal is to predict attribute scores as similar as possible to  ${D}_{g}$. 
The cost function to be minimized is defined as mean squared error:
\begin{align}
C = \frac{1}{M}  \frac{1}{{N}_{w}}  \sum_{i} \sum_{w} [{D}_{g,iw} - {D}_{p,iw}] ^{2}
\end{align}
where  ${\textbf{D}}_{p,i} = [{D}_{p,i1}, {D}_{p,i2}, \dots , {D}_{p,i{N}_{w}}]$ is predictive attribute score vector for $i$th image. $M$ denotes the number of training images.
Convolutional layers are followed by four fully-connected (FC) layers: the first three have 2048 channels each, the fourth contains ${N}_{w}$ channels. 
We use ReLU~\cite{nair2010rectified} as nonlinear activation function for all FC. 
We adopt batch normalization (BN)~\cite{ioffe2015batch} right after each FC and before activation. 
The training is regularized by dropout with ratio 0.3 for the first three FCs. 
Each FC is initialized with a Xavier initialization~\cite{glorot2010understanding}. 
We note that our network does not contain softmax as a final layer, different from other attribute predictors described in previous papers~\cite{gan2017semantic,wu2016value}. 
Hence, we use the output of an activation function of the fourth FC layer as the final predictive score ${\textbf{D}}_{p,i}$.

\section{Experiment}
\label{sec:Experiment}
\subsection{Datasets}

Our results are evaluated on the popular MS COCO dataset~\cite{lin2014microsoft,chen2015microsoft}. The dataset contains 82,783 images for training and 40,504 for validation. 
Due to annotations for test set is not available, we report results with the widely used split ~\cite{karpathy2015deep} which contain 5,000 images for validation and test, respectively.
We applied the same splits to both semantic attribute prediction network and SCN-LSTM network. 
We infer the results of the actual COCO test set consisting of 40,775 images and also evaluate them on the COCO evaluation server~\cite{chen2015microsoft}.

\subsection{Training}
\label{subsec:Training}

The model described in Section~\ref{subsec:model} is used for distinctive-attribute prediction and the training procedures of it are implemented in Keras~\cite{chollet2015keras}.
To implement TF-IDF schemes for meaningful information extraction, we used scikit-learn toolkit~\cite{pedregosa2011scikit}.
The mini-batch size is fixed at $128$ and Adam's optimization~\cite{kingma2014adam} with learning rate $3 \times {10}^{-3}$ is used and stopped after 100 epochs. 
For the prediction model, we train $5$ identical models with different initializations, and then ensemble by averaging their outcomes.
Attributes of training and validation sets are inferred from the prediction model and applied to the SCN-LSTM model training.

In order to analyze the effect of semantic information extraction method on overall performance, various experiments were conducted. 
A vocabulary selection in the semantic information affects training performance, which ultimately affects caption generation performance. 
We use various combinations of vocabularies for the experiment and report both quantitative and qualitative evaluations.
First, we apply IDF thresholding to eliminate the words from vocabulary which have small values than the threshold $th_{IDF}$. We use seven different $th_{IDF}$s for the experiment.
Secondly, we apply stemming for words before extracting TF-IDF and IDF thresholding. After semantic information vectors are extracted, they are fed into the prediction model in pairs with images.
The training results with the different vectors will be reported in Sec~\ref{subsec:results}.

SCN-LSTM training procedure generally follows~\cite{gan2017semantic} except for the dimension of the input attribute vector.
We use the public implementation~\cite{gan2017github} of this method opened by Gan who is the author of the published paper~\cite{gan2017semantic}.
For an image feature, we take out the output of the 2048-way pool5 layer from ResNet-152 which is pre-trained on the ImageNet dataset~\cite{russakovsky2015imagenet}. 
Word embedding vectors are initialized with the word2vec vectors proposed by~\cite{mikolov2013distributed}. 
The number of hidden units and the number of factors are both set to 512. 
We set batch size as 64 and use gradient clipping~\cite{sutskever2014sequence} and dropout~\cite{zaremba2014recurrent}. 
Early stopping was applied for validation sets with the maximum number of epochs 20. 
Adam optimizer~\cite{kingma2014adam} was used with learning rate $2 \times {10}^{-4}$. 
In testing, we use beam search for caption generation and select the top 5 best words at each LSTM step as the candidates. 
We average inferred probability for 5 identical SCN-LSTM model as~\cite{gan2017semantic} did.
    
\subsection{Evaluation Procedures}

We use the macro-average F1 metric to compare the performance of the proposed distinctive-attribute prediction model. 
The output attribute of previous methods~\cite{gan2017semantic,fang2015captions,wu2016value,you2016image} represent probabilities, on the other hand, that of the proposed method are the distinctiveness score itself. 
We evaluate the prediction considering it as a multi-label and multi-class classification problem. The distinctiveness score between 0 and 1 are divided into four classes; $(0.0, 0.25]$, $(0.25, 0.5]$, $(0.5, 0.75]$, and $(0.75, 1.0]$. 
In case the value 0.0 occupies most of the elements, it disturbs accurately comparing the performance. Therefore, we exclude those elements intentionally in the comparison. 
Each word in attribute vocabulary is regarded as one class, respectively. 
The macro-averaged F1 score is computed globally by counting the total number of true positives, false negatives, true negatives, and false positives.

The widely used metrics, BLEU-1,2,3,4~\cite{papineni2002bleu}, METEOR~\cite{banerjee2005meteor}, ROUGL-L~\cite{lin2004rouge}, CIDEr~\cite{vedantam2015cider} are selected to evaluate overall captioning performance. The code released by the COCO evaluation server~\cite{chen2015microsoft} is used for computation.

\subsection{Results}
\label{subsec:results}

Firstly, we compared our method with SCN~\cite{gan2017github} that uses the extracted attribute according to their semantic concept detection method. We evaluate both results on the online COCO testing server and list them in Table~\ref{table:COCO_server}. The pre-trained weights of SCN are provided by the author. We downloaded and used them for an inference according to the author's guide. For the proposed method, we used vocabulary after stemming and set threshold IDF value as 7 in this evaluation. The vocabulary size of the proposed scheme is 938, which is smaller than that of SCN~\cite{gan2017github} with $999$. Accordingly, weight matrices dimensions of the proposed method are smaller than that of SCN in SCN-LSTM structures. Results of both methods are derived from ensembling 5 models, respectively. DaE improves the performance of SCN-LSTM by significant margins across all metrics. Specifically, DaE improves CIDEr from $0.967$ to $0.981$ in 5-refs and from $0.971$ to $0.990$ in 40-refs. The increase is greater at 40-refs. The Proposed method can be applied to other base models that use attributes to improve their performance. The results for other published models tested on the COCO evaluation server are summarized in Table~\ref{table:COCO_server_published}. In 40-refs, our method surpasses the performance of $Addaptive Attention + CL$ which is the state-of-the-art in terms of four BLEU scores. 
\setlength{\tabcolsep}{4pt}
\begin{table}
	\begin{center}
		\caption{COCO evaluation server results using 5 references and 40 references captions. BLEU-1,2,3,4, METEOR, ROUGE-L, CIDEr metrics are used to comparing SCN and the proposed method. DaE improves the performance by significant margins across all metrics}
		\label{table:COCO_server}
		\begin{footnotesize}
			\begin{tabularx}{\textwidth}{lXXXXXXX}
				\hline
				& B-1 & B-2 & B-3 & B-4 & M & R & CIDEr\\
				\hline
				\multicolumn{8}{l}{5-refs} \\
				\hline
				SCN & 0.729 & 0.563 & 0.426 & \bf0.324 & 0.253 & 0.537 & 0.967 \\
				DaE + SCN-LSTM & \bf0.734 & \bf0.568 & \bf0.429 & \bf0.324 & \bf0.255 & \bf0.538 & \bf0.981 \\
				\hline
				\multicolumn{8}{l}{40-refs} \\
				\hline
				SCN & 0.910 & 0.829 & 0.727 & 0.619 & 0.344 & 0.690 & 0.971 \\
				DaE + SCN-LSTM & \bf0.916 & \bf0.836 & \bf0.734 & \bf0.625 & \bf0.348 & \bf0.694 & \bf0.990 \\
				\hline
			\end{tabularx}
		\end{footnotesize}
	\end{center}
\end{table}
\setlength{\tabcolsep}{1.5pt}
\setlength{\tabcolsep}{4pt}
\begin{table}
\begin{center}
\caption{Results of published image captioning models tested on the COCO evaluation server.}
\label{table:COCO_server_published}
\begin{scriptsize}
	\begin{tabularx}{\textwidth}{lXXXXXXX}
		\hline
		& B-1 & B-2 & B-3 & B-4 & M & R & CIDEr\\
		\hline
		\multicolumn{8}{l}{5-refs} \\
		\hline
		Hard-Attention~\cite{xu2015show} & 0.705 & 0.528 & 0.383 & 0.277 & 0.241 & 0.516 & 0.865 \\
		Google NIC~\cite{vinyals2015show} & 0.713 & 0.542 & 0.407 & 0.309 & 0.254 & 0.530 & 0.943 \\
		ATT-FCN~\cite{you2016image} & 0.731 & 0.565 & 0.424 & 0.316 & 0.250 & 0.535 & 0.943 \\
		Adaptive Attention~\cite{lu2017knowing} & 0.735 & 0.569 & 0.429 & 0.323 & 0.258 & 0.541 & 1.001 \\
		Addaptive Attention + CL~\cite{dai2017contrastive} & 0.742 & 0.577 & 0.436 & 0.326 & 0.260 & 0.544 & 1.010 \\
		DaE + SCN-LSTM & 0.734 & 0.568 & 0.429 & 0.324 & 0.255 & 0.538 & 0.981 \\
		\hline
		\multicolumn{8}{l}{40-refs} \\
		\hline
		Hard-Attention~\cite{xu2015show} & 0.881 & 0.779 & 0.658 & 0.537 & 0.322 & 0.654 & 0.893 \\
		Google NIC~\cite{vinyals2015show} & 0.895 & 0.802 & 0.694 & 0.587 & 0.346 & 0.682 & 0.946 \\
		ATT-FCN~\cite{you2016image} & 0.900 & 0.815 & 0.709 & 0.599 & 0.335 & 0.682 & 0.958 \\
		Adaptive Attention~\cite{lu2017knowing} & 0.906 & 0.823 & 0.717 & 0.607 & 0.347 & 0.689 & 1.004 \\
		Addaptive Attention + CL~\cite{dai2017contrastive} & 0.910 & 0.831 & 0.728 & 0.617 & \bf0.350 & \bf0.695 & \bf1.029 \\
		DaE + SCN-LSTM & \bf0.916 & \bf0.836 & \bf0.734 & \bf0.625 & 0.348 & 0.694 & 0.990 \\
		\hline
	\end{tabularx}
\end{scriptsize}
\end{center}
\end{table}
\setlength{\tabcolsep}{1.5pt}

For the qualitative evaluation, tags extracted by the semantic concept detection of the SCN and  description generated using them are illustrated as shown in Table~\ref{table:server_qualitative}. Moreover, distinctive-attributes extracted by DaE and a caption are shown in the lower row.
The attributes extracted using DaE include important words to represent the situation in an image; as a result, the caption generated by using them are represented more in detail compared with those of SCN.
Scores in the right parentheses of the tags and distinctive-attributes have different meanings, the former is probabilities, and the latter is distinctiveness values of words by the proposed scheme. We listed the top eight attributes in descending order.
In the case of DaE, words after stemming with Porter Stemmer~\cite{porter1980algorithm} are displayed as they are.
The result of OURS in (a), ``A woman cutting a piece of fruit with a knife", explains exactly what the main character does.
In the SCN, the general word `food' get a high probability,  on the other hand, DaE extracts more distinctive words such as `fruit' and ``apple". 
For verbs, ``cut", which is the most specific action that viewers would be interested in, gets high distinctiveness score. 
In the case of (b), ``wine" and ``drink" are chosen as the words with the first and the third highest distinctiveness through DaE. Therefore, the characteristic phrase ``drinking wine" is added.
\setlength{\tabcolsep}{4pt}
\begin{table}
	\begin{center}
		\caption{This figure illustrates several images with extracted attributes and captions. For attribute extraction, SCN~\cite{gan2017semantic} uses their semantic concept detection method, and ours uses DaE. Both use SCN-LSTM to generate captions. The captions generated by using DaE+SCN-LSTM are explained more in detail with more distinctive and accurate attributes}
		\label{table:server_qualitative}
		\begin{tabularx}{\textwidth}{|c|X|X|X|}
			\hline
			&
			\multicolumn{1}{c|}{\tiny (a)}& 
			\multicolumn{1}{c|}{\tiny (b)}&  
			\multicolumn{1}{c|}{\tiny (c)} \\
			\hline
			&
			\centering
			\includegraphics[width=1.8cm, height=2.1cm]{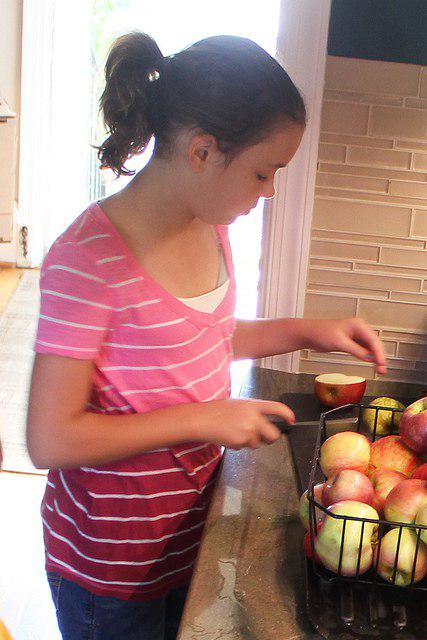} & 
			\includegraphics[width=3.5cm, height=2.1cm]{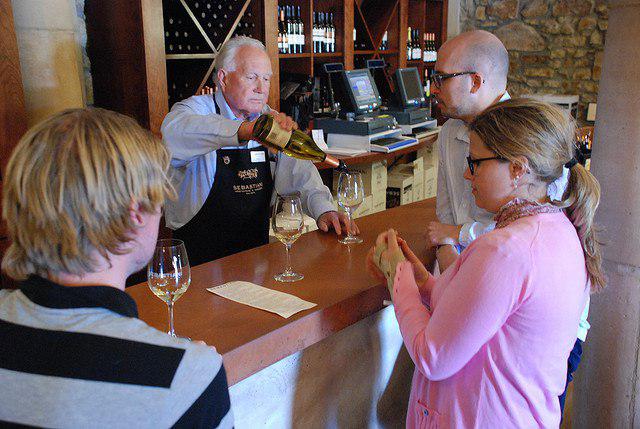} &
			\includegraphics[width=3.5cm, height=2.1cm]{images/server_ex3.jpg} \\
			\hline
			{\tiny SCN} & 
			{\tiny \begin{tabular}{@{}p{3.5cm}@{}}\\Generated captions:\\\bf A woman standing in a kitchen preparing food\\ \\ Tags:\\ person (0.99), food (0.91), indoor (0.85), table (0.58), woman (0.51), preparing (0.50), kitchen (0.42), small (0.35)\\ \\ \end{tabular}} & 
			
			{\tiny \begin{tabular}{@{}p{3.5cm}@{}}\\Generated captions:\\\bf A group of people sitting at a table\\ \\ Tags:\\ person (1.00), table (0.99), indoor (0.90), sitting (0.80), woman (0.76), man (0.57), front (0.46), group (0.36)\\ \\\end{tabular}} & 
			
			{\tiny \begin{tabular}{@{}p{3.5cm}@{}}\\Generated captions:\\\bf A group of people standing in front of a table\\ \\ Tags:\\ indoor (0.83), table (0.63), standing (0.51), photo (0.49), computer (0.34), front (0.31), man (0.31), next (0.26)\\ \\\end{tabular}} \\
			\hline
			
			{\tiny \begin{tabular}{@{}m{0.6cm}@{}}\\DaE\\\\+\\\\SCN-\\LSTM\end{tabular}} & 
			
			{\tiny \begin{tabular}{@{}p{3.5cm}@{}}\\Generated captions:\\ \bf A woman cutting a piece of fruit with a knife\\ \\ Distinctive-attribute:\\ \textbf{cut} (0.41), woman (0.28), \textbf{knife} (0.27), cake (0.18), \textbf{fruit} (0.14), food (0.13), kitchen (0.42), \textbf{appl} (0.11)\\ \\\end{tabular}} & 
			
			{\tiny \begin{tabular}{@{}p{3.5cm}@{}}\\Generated captions:\\\bf A group of people sitting at a table drinking wine\\ \\ Distinctive-attribute:\\ \textbf{wine} (0.41), peopl (0.16), \textbf{drink} (0.13), tabl (0.12), woman (0.09), man (0.07), girl (0.07), group (0.06)\\ \\\end{tabular}} & 
			
			{\tiny \begin{tabular}{@{}p{3.5cm}@{}}\\Generated captions:\\\bf A room filled with lots of colorful decorations\\ \\ Distinctive-attribute:\\ \textbf{color} (0.15), room (0.12), \textbf{decor} (0.11), hang (0.10), display (0.10), of (0.09), with (0.08), and (0.07)\\ \\\end{tabular}} \\
			\hline
			&
			\multicolumn{1}{c|}{\tiny (d)}& 
			\multicolumn{1}{c|}{\tiny (e)}&  
			\multicolumn{1}{c|}{\tiny (f)} \\
			\hline
			&
			\centering
			\includegraphics[width=3.5cm, height=2.1cm]{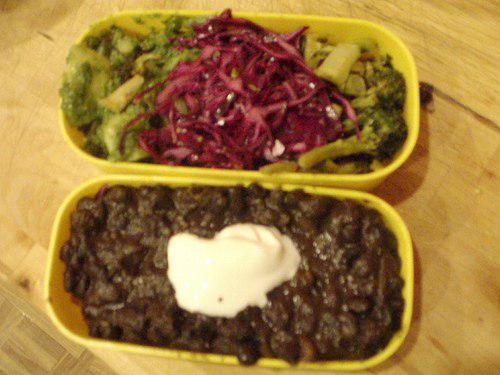} & 
			\includegraphics[width=3.5cm, height=2.1cm]{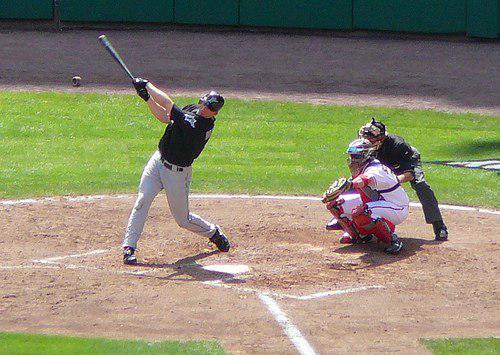} &
			\includegraphics[width=3.5cm, height=2.1cm]{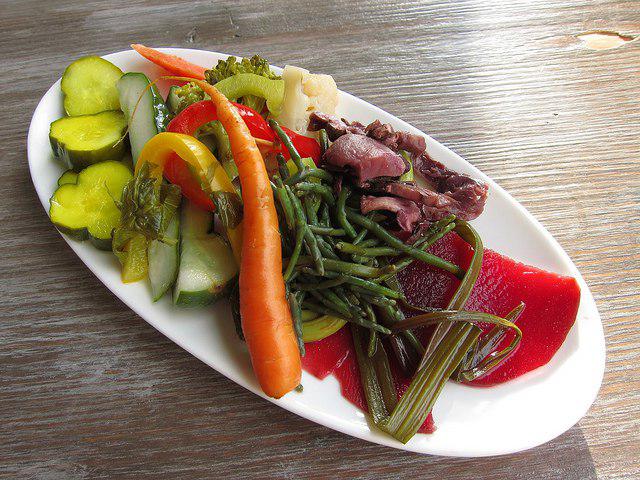} \\
			\hline
			{\tiny SCN} & 
			{\tiny \begin{tabular}{@{}p{3.5cm}@{}}\\Generated captions:\\\bf A close up of a bowl of food\\ \\ Tags:\\ food (1.00), table (0.97), indoor (0.92), container (0.71), sitting (0.67), wooden (0.61), sauce (0.53), plate (0.53)\\ \\ \end{tabular}} & 
			
			{\tiny \begin{tabular}{@{}p{3.5cm}@{}}\\Generated captions:\\\bf A baseball player swinging a bat at a ball\\ \\ Tags:\\ grass (1.00), baseball (1.00), player (0.99), bat (0.97), person (0.95), game (0.95), sport (0.95), swinging (0.93)\\ \\\end{tabular}} & 
			
			{\tiny \begin{tabular}{@{}p{3.5cm}@{}}\\Generated captions:\\\bf A close up of a plate of food on a table\\ \\ Tags:\\ food (1.00), plate (0.99), table (0.98), hot (0.43), sitting (0.35), small (0.29), fruit (0.24), filled (0.23)\\ \\\end{tabular}} \\
			\hline
			
			{\tiny \begin{tabular}{@{}m{0.6cm}@{}}\\DaE\\\\+\\\\SCN-\\LSTM\end{tabular}} & 
			
			{\tiny \begin{tabular}{@{}p{3.5cm}@{}}\\Generated captions:\\ \bf Two plastic containers filled with different types of food\\ \\ Distinctive-attribute:\\ contain (0.34), food (0.22), \textbf{veget} (0.16), \textbf{and} (0.12), \textbf{brocoli} (0.11), dish (0.09), \textbf{meat (0.08)}, of (0.09)\\ \\\end{tabular}} & 
			
			{\tiny \begin{tabular}{@{}p{3.5cm}@{}}\\Generated captions:\\\bf A batter catcher and umpire during a baseball game\\ \\ Distinctive-attribute:\\ basebal (0.49), bat (0.32), player (0.18), swing (0.18), \textbf{catcher} (0.11), \textbf{umpir} (0.11), ball (0.10), \textbf{batter} (0.10)\\ \\\end{tabular}} & 
			
			{\tiny \begin{tabular}{@{}p{3.5cm}@{}}\\Generated captions:\\\bf A white plate topped with a variety of vegetables\\ \\ Distinctive-attribute:\\ plate (0.48), \textbf{veget} (0.33), \textbf{carrot} (0.16), \textbf{salad} (0.16), \textbf{and} (0.13), food (0.10), on (0.09), with (0.09)\\ \\\end{tabular}} \\
			\hline
			
		\end{tabularx}
	\end{center}
\end{table}
\setlength{\tabcolsep}{1.5pt}

To analyze DaE in more detail, we conduct experiments with differently constructed vocabularies, as explained in Section~\ref{subsec:Training}. 
We used splits on COCO training and validation sets as done in the work of~\cite{karpathy2015deep}.
Table~\ref{table:DaE_stem}(a) presents the results of experiments with vocabularies after stemming.
We set seven different IDF threshold values, ${th}_{IDF}$, from 5 to 11. 
\begin{align}
{Vocab}_{i} \; (i \in \{5,6, \dots , 11\}),
\; \\
{Vocab}_{i} = \{ w \;| \; IDF(w) > i,  i = {th}_{IDF}\}.
\end{align}
The vocabulary contains only the words whose IDF is bigger than ${th}_{IDF}$.
Setting the IDF threshold value to 5 means that only the words appearing in over $1/{10}^4$ of the entire images are treated, according to \ref{eq:IDF}.
The number of vocabulary words is shown in the second row of Table~\ref{table:DaE_stem}(a). For example, the number of words in ${Vocab}_{5}$ is 276 out of total 5,663 words after stemming in reference captions.
Semantic information of the images are extracted corresponding to this vocabulary, and we use them to learn the proposed prediction model.
The performance, macro-averaged F1, of the prediction model evaluated by test splits is shown in the third row.
The lower the ${th}_{IDF}$, that is, the vocabulary is composed of the more frequent words, provides the better prediction performance. 
Each extracted distinctive-attribute is fed into SCN-LSTM to generate a caption, and the evaluation result, CIDEr, is shown in the fourth row.
The CIDErs increase from ${Vocab}_{5}$ to ${Vocab}_{7}$, and then monotonically decrease in the rest. In other words, the maximum performance is derived from ${Vocab}_{7}$ to 0.996.
The vocabulary size and the prediction performance are in a trade-off in this experiment. With the high ${th}_{IDF}$ value, captions can be generated with various vocabularies, but the captioning performance is not maximized because the performance of distinctive-attribute prediction is relatively low. 
\setlength{\tabcolsep}{4pt}
\begin{table}
\begin{center}
\caption{Results of experiments with differently constructed vocabularies with stemming. (a) and (b) represent results with stemming and without stemming, respectively. The prediction performance in F1 decreases from ${Vocab}_{5}$ to ${Vocab}_{11}$, and the best captioning performance in CIDEr is obtained at ${Vocab}_{7}$ in both (a) and (b)}
		\label{table:DaE_stem}
		\begin{tabularx}{\textwidth}{lXXXXXXX}
			\hline
			& Vocab$_{5}$ & Vocab$_{6}$ & Vocab$_{7}$ & Vocab$_{8}$ & Vocab$_{9}$ & Vocab$_{10}$ & Vocab$_{11}$\\
			\hline
			\hline
			\textbf{(a) With stemming} \\
			\hline
			\# of vocabulary & 276 & 546 & 938 & 1660 & 2656 & 4009 & \textbf{5530}\\
			\hline
			F1(DaE) &\textbf{0.432} & 0.401 & 0.389 & 0.379 & 0.378 & 0.373 & 0.374\\
			\hline
			CIDEr(caption) & 0.978 & 0.991 & \textbf{0.996} & 0.994 & 0.991 & 0.984 & 0.981\\
			\hline
			\hline
			\textbf{(b) Without stemming} \\
			\hline
			\# of vocabulary & 241 & 582 & 1121 & 2039 & 3572 & 5900 & \textbf{8609}\\
			\hline
			F1(DaE) &\textbf{0.437} & 0.399 & 0.383 & 0.374 & 0.366 & 0.362 & 0.358\\
			\hline
			CIDEr(caption) & 0.955 & 0.989 & \textbf{0.991} & 0.986 & 0.990 & 0.988 & 0.979\\
			\hline
		\end{tabularx}
	\end{center}
\end{table}
\setlength{\tabcolsep}{1.5pt}

\setlength{\tabcolsep}{4pt}
\begin{table}
	\begin{center}
		\caption{Several cases that more diverse and accurate captions are generated using ${Vocab}_{9}$ than using ${Vocab}_{6}$, although their CIDErs are similar}
		\label{table:DaE_qualitative}
		\begin{tabularx}{\textwidth}{|c|X|X|X|}
			\hline
			&
			\multicolumn{1}{c|}{\tiny (a)}& 
			\multicolumn{1}{c|}{\tiny (b)}&  
			\multicolumn{1}{c|}{\tiny (c)} \\
			\hline
			&
			\centering
			\includegraphics[width=3.5cm, height=2.1cm]{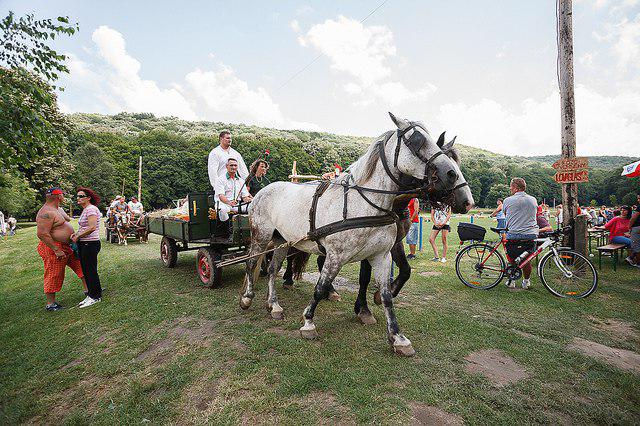} & 
			\includegraphics[width=3.5cm, height=2.1cm]{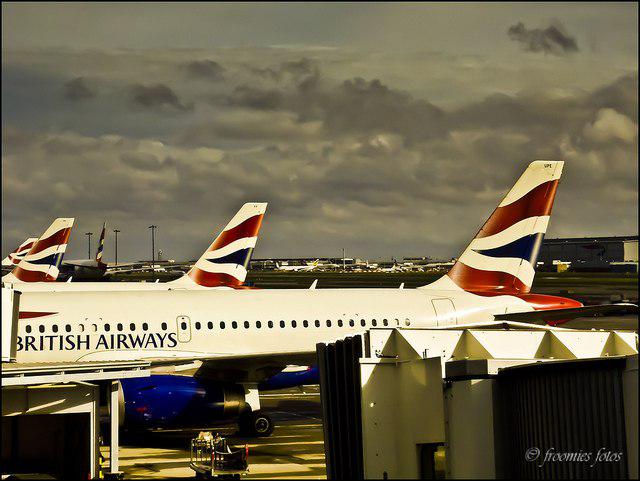} &
			\includegraphics[width=3.5cm, height=2.1cm]{images/dae_ex3.jpg} \\
			\hline
			{\tiny \begin{tabular}{@{}m{0.8cm}@{}}\\Vocab$_{6}$\end{tabular}}  & 
			{\tiny \begin{tabular}{@{}p{3.4cm}@{}}\\Generated captions:\\\bf A couple of people standing next to a horse\\ \\ Distinctive-attribute:\\ hors (0.58), pull (0.11), peopl (0.10), two (0.10), stand (0.07), field (0.07), of (0.07), in (0.06)\\ \\ \end{tabular}} & 
			
			{\tiny \begin{tabular}{@{}p{3.4cm}@{}}\\Generated captions:\\\bf A large air plane on a run way\\ \\ Distinctive-attribute:\\ airport (0.28), plane (0.26), airplan (0.25), jet (0.22), park (0.13), runway (0.12), an (0.12), on (0.09)\\ \\\end{tabular}} & 
			
			{\tiny \begin{tabular}{@{}p{3.4cm}@{}}\\Generated captions:\\\bf A toaster oven sitting on top of a counter\\ \\ Distinctive-attribute:\\ oven (0.51), counter (0.18), kitchen (0.13), on (0.06), of (0.06), top (0.06), an (0.05), in (0.05)\\ \\\end{tabular}} \\
			\hline
			
			{\tiny \begin{tabular}{@{}m{0.8cm}@{}}\\Vocab$_{9}$\end{tabular}} & 
			
			{\tiny \begin{tabular}{@{}p{3.4cm}@{}}\\Generated captions:\\\bf A couple of horses pulling a carriage in a field\\ \\ Distinctive-attribute:\\ horse (0.58), pull (0.17), peopl (0.10), two (0.08), of (0.07), in (0.07), \textbf{carriag} (0.06), stand (0.06)\\ 
					\\\end{tabular}} & 
			
			{\tiny \begin{tabular}{@{}p{3.4cm}@{}}\\Generated captions:\\\bf A large jetliner sitting on top of an airport tarmac\\ \\ Distinctive-attribute:\\ airport (0.30), airplan (0.28), plane (0.25), jet (0.18), runway (0.16), an (0.12), \textbf{tarmac} (0.12), park (0.09)\\ \\\end{tabular}} & 
			
			{\tiny \begin{tabular}{@{}p{3.4cm}@{}}\\Generated captions:\\\bf A microwave oven sitting on top of a counter\\ \\ Distinctive-attribute:\\ oven (0.46), \textbf{microwav} (0.40), counter (0.14), kitchen (0.07), on (0.06), of (0.06), top (0.06), an (0.05)\\ \\\end{tabular}} \\
			\hline
			
		\end{tabularx} 
	\end{center}
\end{table}
\setlength{\tabcolsep}{1.5pt}

${Vocab}_{6}$ and ${Vocab}_{9}$ have almost the same CIDEr. At this time, If the vocabulary contains more words, it is possible to represent the captions more diversely and accurately for some images. Table~\ref{table:DaE_qualitative} shows examples corresponding to this case. For the case of (a), the ${Vocab}_{6}$ does not include the word ``carriag", but the ${Vocab}_{9}$ contains the words and is extracted as the word having the seventh highest value through DaE. This led the phrase ``pulling a carriage" to be included the caption, well describing the situation. ``Tamac" in (b), and ``microwav" in (c) plays a similar role.

Table~\ref{table:DaE_stem} (b) presents experimental results without stemming. The captioning performance is highest at ${Vocab}_{7}$. 
The value was 0.911, which is lower than the maximum value of the experiments with stemming.
When stemming is applied, the distinctiveness and significance of a word can be better expressed because it is mapped to the same word even if the tense and form are different.
The size of vocabulary required to achieve the same performance is less when stemming is applied. 
It means that the number of parameters needed for the captioning model is small and the computational complexity is low.

\section{Conclusion}
\label{sec:conclusion}
In this study, we propose a Distinctive-attribute Extraction (DaE) method for image captioning. In particular, the proposed scheme consists of the semantic attribute extraction and semantic attribute prediction. 
To obtain the semantic attributes, TF-IDF of trained captions is computed to extract meaningful information from them. Then, the distinctive-attribute vectors for an image are computed by regularizing TF-IDF of each word with the $L2$ normalized TF-IDF of the image. The attribute prediction model is trained by the extracted attributes and used to infer the semantic-attribute for generating a natural language description. 
DaE improves the performance of SCN-LSTM scheme by signicant margins across all metrics, moreover, distinctive captions are generated. Specifically, CIDEr scores on the COCO evaluation server are improved from 0.967 to 0.981 in 5-refs and from 0.971 to 0.990
in 40-refs, respectively. The proposed method can be applied to other base models that use attribute to improve their performance.
Therefore, we believe that the proposed scheme can be a useful tool for effective image caption scheme.

\clearpage   
                           
\section*{Supplementary Material}	

\vspace{0.5cm}
	
	\noindent In the experiment, we compared our method with SCN~\cite{gan2017semantic,gan2017github} that uses extracted tags according to their semantic concept detection method. To evaluate the proposed method with more pictures, we compare the predicted semantic attributes by using SCN and the proposed scheme. The results are listed in Table~\ref{table:table}.
	The attribute in SCN and the proposed method (DaE) is called as tag and distinctive-attribute, respectively.
	The tag represents probabilities, on the other hand, the attribute from DaE is distinctiveness score itself.
	We listed the top eight attributes in descending order. In the case of DaE, words after stemming are displayed as they are.
	The captions obtained using image features and extracted semantic information are also compared in the table.
	\\\\
	In (a), a child is feeding grass to a giraffe through a fence. The caption generated by SCN includes ``dog"  that does not exist in the picture and is inaccurate. However, as a result of DaE, the word ``giraff" gets a higher score than the ``dog" and is reflected in the generated caption. In addition, DaE detects the verb ``feed", which represents the main situation of the image, and the exact phrase ``feeding a giraffe through a fence" is produced. 
	\\\\
	In (b), ``red truck" and ``snow" are recognized as ``fire hydrant" and ``water," respectively, by SCN. Those words creating the phrase ``hydrant spraying water" that does not fit a situation of the image. On the other hand, DaE extracts exact nouns, verb and adjective such as ``truck", ``snow," ``drive," and ``red."
	\\\\
	In (c), DaE detects the banana located in a small part of the image with the highest score among the distinctive-attributes. ``Banana" is combined with another well-detected word ``hold" to create a participial construction: ``holding a banana."
	\\\\
	In (d), the situation is that a man is taking selfi through a mirror. DaE detects the stemmed word ``hi" corresponding to ``himself." On the other hand, the tag vocabulary set of SCN does not contain the words such as ``himself" or ``self." Besides, SCN recognizes the camera or phone as a Nintendo.
	\\\\
	In (e), the general caption ``A close up of a sandwich on a plate." is generated by SCN, on the other hand, the caption generated using the proposed method contains a distinctive phrase ``cut in half" due to the extracted distinctive-attributes ``cut" and ``half."
	\\\\
	In (f), there is a bull in the center of the picture. The vocabulary of SCN does not contain the word ``bull", but the vocabulary of our method contains the word, even though the vocabulary size is smaller. This specific word is extracted through DaE and reflected in the caption.
	\\\\
	In (g), DaE detects that the picture is a ``store" or a ``shop," and accurately figures out the situation that the clock is ``displayed" over the ``window." On the other hand, SCN extracts words that are general and inappropriate to the situation, such as ``building" and ``outdoor."
	\\\\
	In (h), there is a red stop sign next to a man. DaE extracts both ``sign" and its message ``stop." In addition, ``sunglass" is extracted to generate a caption that well represents an appearance of the man. On the other hand, the caption generated by SCN includes expressions such as ``man in a blue shirt" and ``holding a sign" that is not the situation of the picture.
	\\\\
	In (i), DaE extracts the word ``frost" that exists only in its vocabulary and does not exist in the vocabulary of SCN. And the elaborate caption was created containing the word. The caption "A close up of a cake on a plate," which is generated by SCN, is relatively general.
	\\\\
	In (j), DaE extracts key objects and place such as ``microwav", ``kitchen", ``sink", etc. And the captions generated by them are more detailed than captions generated by the tags of SCN.
	\\\\
	In (k), a man is standing in front of a computer monitor or laptops. DaE detects ``comput" and  ``laptop," which are not detected by SCN, and generates more accurate caption than that using the tags of SCN.
	\\\\
	In (l), a pair of scissors placed in a plastic packing case is taken close up.
	DaE extracts ``scissor" which is the main object of the picture as the highest score. The word ``pair" which is used when counting the scissor, is extracted as the second highest score. On the other hand, the main object of the caption generated by SCN is ``cell phone" that does not exist in the picture. 
	
	\setlength{\tabcolsep}{4pt}
	\begin{table}
		\label{table:table}
		\begin{center}
			\caption{This figure illustrates several images with extracted attributes and captions. For attribute extraction, SCN uses their semantic concept detection method, and ours uses DaE. Both uses SCN-LSTM to generate captions. The captions generated by using DaE+SCN-LSTM are explained more in detail with more distinctive and accurate attributes.}
			\label{table:server_qualitative}
			\begin{tabularx}{\textwidth}{|c|X|X|X|}
				\hline
				&
				\multicolumn{1}{c|}{\tiny (a)}& 
				\multicolumn{1}{c|}{\tiny (b)}&  
				\multicolumn{1}{c|}{\tiny (c)} \\
				\hline
				&
				\centering
				\includegraphics[width=3.5cm, height=2.2cm]{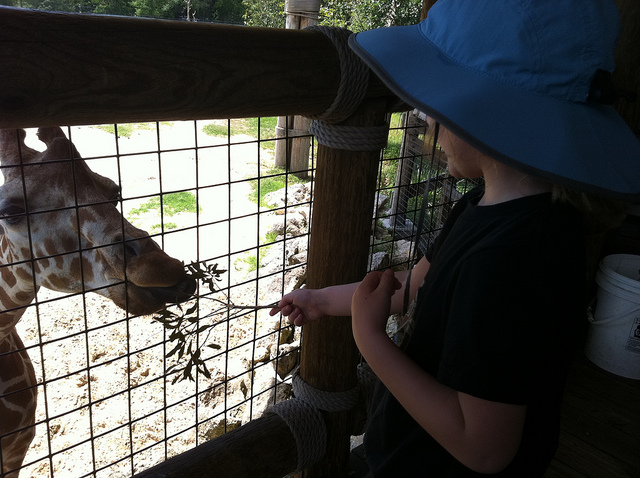} & 
				\includegraphics[width=3.5cm, height=2.2cm]{images/sm2.jpg} &
				\includegraphics[width=3.5cm, height=2.2cm]{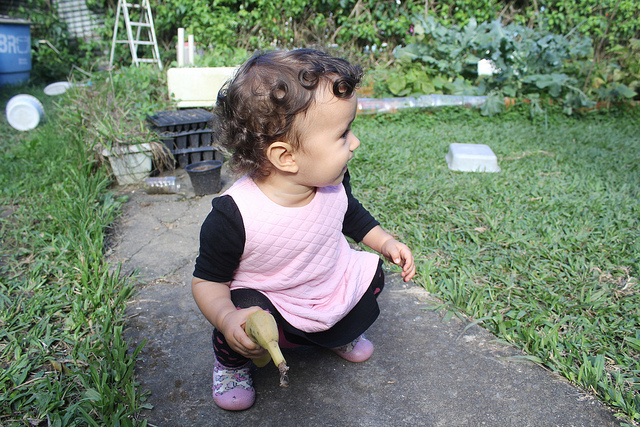} \\
				\hline
				{\tiny SCN} & 
				{\tiny \begin{tabular}{@{}p{3.5cm}@{}}\\Generated captions:\\\bf A dog is looking out of a fence\\ \\ Tags:\\ person (0.99), fence (0.87), building (0.65), window (0.61), looking (0.52), dog (0.47), standing (0.45), small (0.35)\\ \\ \end{tabular}} & 
				
				{\tiny \begin{tabular}{@{}p{3.5cm}@{}}\\Generated captions:\\\bf A fire hydrant spraying water from a fire hydrant\\ \\ Tags:\\ outdoor (0.99), orange (0.97), fire (0.83), water (0.76), hydrant (0.55), car (0.52), yellow (0.46), truck (0.44)\\ \\\end{tabular}} & 
				
				{\tiny \begin{tabular}{@{}p{3.5cm}@{}}\\Generated captions:\\\bf A little boy is playing with a frisbee'\\ \\ Tags:\\ outdoor (1.00), grass (1.00), person (0.99), child (0.98), little (0.97), young (0.94), boy (0.93), small (0.85)\\ \\\end{tabular}} \\
				\hline
				
				{\tiny \begin{tabular}{@{}m{0.6cm}@{}}\\DaE\\\\+\\\\SCN-\\LSTM\end{tabular}} & 
				
				{\tiny \begin{tabular}{@{}p{3.5cm}@{}}\\Generated captions:\\ \bf A person feeding a giraffe through a fence\\ \\ Distinctive-attribute:\\ \textbf{giraff} (0.40), fenc (0.25), \textbf{feed} (0.12), dog (0.10), out (0.07), look (0.06), in (0.05), is (0.05)\\ \\\end{tabular}} & 
				
				{\tiny \begin{tabular}{@{}p{3.5cm}@{}}\\Generated captions:\\\bf A red truck driving down a snow covered road\\ \\ Distinctive-attribute:\\ \textbf{truck} (0.40), \textbf{snow} (0.19), orang (0.12), \textbf{drive} (0.11), car (0.09), the (0.09), toy (0.08), \textbf{red} (0.07)\\ \\\end{tabular}} & 
				
				{\tiny \begin{tabular}{@{}p{3.5cm}@{}}\\Generated captions:\\\bf A small child sitting on the ground holding a banana\\ \\ Distinctive-attribute:\\ \textbf{banana} (0.35), boy (0.22), child (0.18), little (0.15), \textbf{hold} (0.12), young (0.11), skateboard (0.09), on (0.08)\\ \\\end{tabular}} \\
				\hline
				&
				\multicolumn{1}{c|}{\tiny (d)}& 
				\multicolumn{1}{c|}{\tiny (e)}&  
				\multicolumn{1}{c|}{\tiny (f)} \\
				\hline
				&
				\centering
				\includegraphics[width=3.5cm, height=2.2cm]{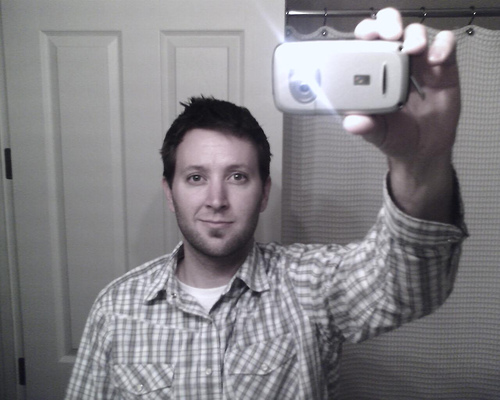} & 
				\includegraphics[width=3.5cm, height=2.2cm]{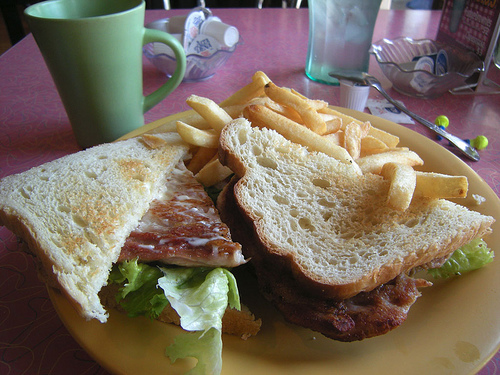} &
				\includegraphics[width=3.5cm, height=2.2cm]{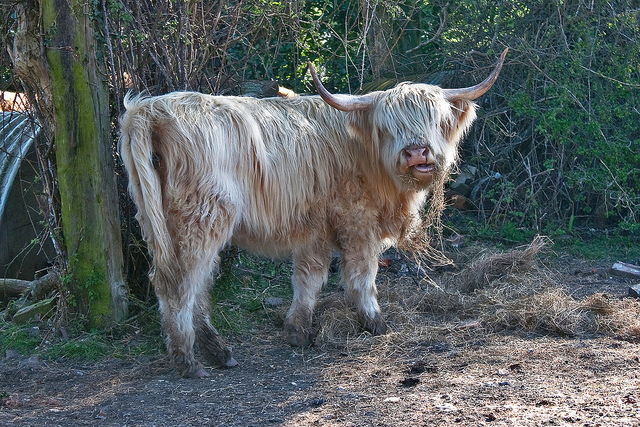} \\
				\hline
				{\tiny SCN} & 
				{\tiny \begin{tabular}{@{}p{3.5cm}@{}}\\Generated captions:\\\bf A man holding a nintendo wii game controller\ \\ Tags:\\ person (1.0), indoor (0.99), holding (0.99), man (0.96), controller (0.91), remote (0.89), video (0.87)\\ \\ \end{tabular}} & 
				
				{\tiny \begin{tabular}{@{}p{3.5cm}@{}}\\Generated captions:\\\bf A close up of a sandwich on a plate\\ \\ Tags:\\ food (1.00), sandwich (1.00), cup (0.98), plate (0.94), dish (0.90), indoor (0.87), sitting (0.84), coffee (0.80)\\ \\ \end{tabular}} &
				
				{\tiny \begin{tabular}{@{}p{3.5cm}@{}}\\Generated captions:\\\bf A close up of a cow in a field\\ \\ Tags:\\ outdoor (1.00), grass (0.97), cow (0.97), animal (0.95), mammal (0.93), standing (0.87), hay (0.79), brown (0.64)\\ \\\end{tabular}} \\
				\hline
				
				{\tiny \begin{tabular}{@{}m{0.6cm}@{}}\\DaE\\\\+\\\\SCN-\\LSTM\end{tabular}} & 
				
				{\tiny \begin{tabular}{@{}p{3.5cm}@{}}\\Generated captions:\\ \bf A man is taking a picture of himself\\ \\ Distinctive-attribute:\\ \textbf{take} (0.35), man (0.27), \textbf{phone} (0.24), hold (0.20), \textbf{hi} (0.19), pictur (0.17), \textbf{camera} (0.15), \textbf{cell} (0.14)\\ \\\end{tabular}} & 
				
				{\tiny \begin{tabular}{@{}p{3.5cm}@{}}\\Generated captions:\\\bf A sandwich cut in half on a plate\\ \\ Distinctive-attribute:\\ sandwich (0.70), plate (0.28), \textbf{cut} (0.16), \textbf{half} (0.13), and (0.11), on (0.10), with (0.09), fri (0.09)\\ \\\end{tabular}} &
				
				{\tiny \begin{tabular}{@{}p{3.5cm}@{}}\\Generated captions:\\\bf A bull is standing next to a tree\\ \\ Distinctive-attribute:\\ cow (0.27), stand (0.19), tree (0.13), in (0.09), \textbf{bull} (0.08), brown (0.08), the (0.06), field (0.06)\\ \\\end{tabular}} \\
				\hline
				
			\end{tabularx}
		\end{center}
	\end{table}
	\setlength{\tabcolsep}{1.5pt}

	\setlength{\tabcolsep}{4pt}
	\begin{table}
		\begin{center}
			\begin{tabularx}{\textwidth}{|c|X|X|X|}
				\hline
				&
				\multicolumn{1}{c|}{\tiny (g)}& 
				\multicolumn{1}{c|}{\tiny (h)}&  
				\multicolumn{1}{c|}{\tiny (i)} \\
				\hline
				&
				\centering
				\includegraphics[width=3.5cm, height=2.2cm]{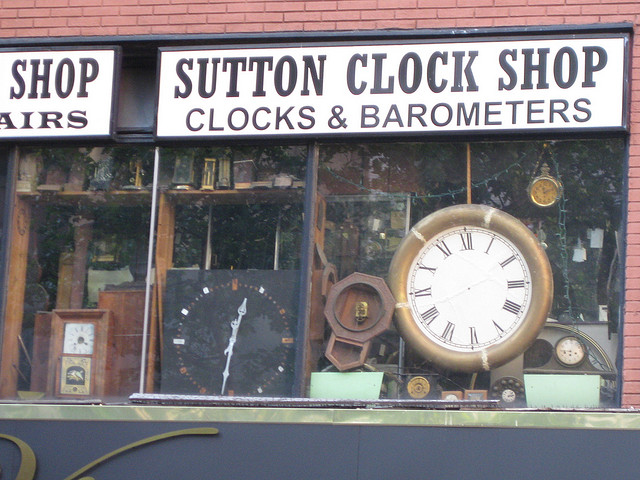} & 
				\includegraphics[width=3.5cm, height=2.2cm]{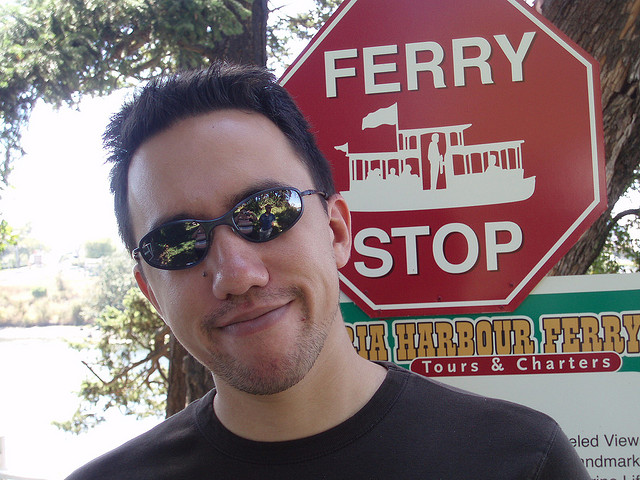} &
				\includegraphics[width=3.5cm, height=2.2cm]{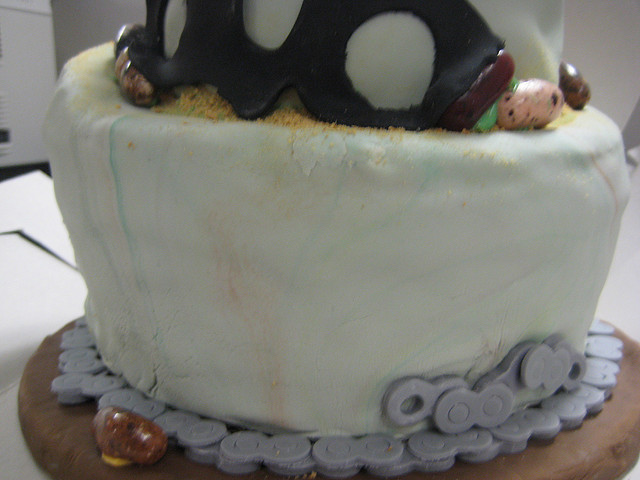} \\
				\hline
				{\tiny SCN} & 
				{\tiny \begin{tabular}{@{}p{3.5cm}@{}}\\Generated captions:\\\bf A large clock on the side of a building\\ \\ Tags:\\ building (0.99), outdoor (0.93), clock (0.85), front (0.70), sign (0.49), large (0.44), sitting (0.27), next (0.24)\\ \\ \end{tabular}} & 
				
				{\tiny \begin{tabular}{@{}p{3.5cm}@{}}\\Generated captions:\\\bf A man in a blue shirt is holding a sign\\ \\ Tags:\\ person (1.00), outdoor (1.00), man (0.99), sign (0.65), front (0.61), eating (0.55), holding (0.55), food (0.45)\\ \\\end{tabular}} & 
				
				{\tiny \begin{tabular}{@{}p{3.5cm}@{}}\\Generated captions:\\\bf A close up of a cake on a plate\\ \\ Tags:\\ cake (1.00), food (0.96), plate (0.92), table (0.91), chocolate (0.86), indoor (0.86), decorated (0.85), top (0.83)\\ \\\end{tabular}} \\
				\hline
				
				{\tiny \begin{tabular}{@{}m{0.6cm}@{}}\\DaE\\\\+\\\\SCN-\\LSTM\end{tabular}} & 
				
				{\tiny \begin{tabular}{@{}p{3.5cm}@{}}\\Generated captions:\\ \bf A store window with a clock on display\\ \\ Distinctive-attribute:\\ \textbf{store} (0.33), \textbf{window} (0.32), clock (0.31), \textbf{display} (0.30), \textbf{shop} (0.16), sign (0.10), of (0.09), front (0.07)\\ \\\end{tabular}} & 
				
				{\tiny \begin{tabular}{@{}p{3.5cm}@{}}\\Generated captions:\\\bf A man wearing sunglasses standing next to a stop sign\\ \\ Distinctive-attribute:\\ \textbf{sign} (0.39), \textbf{stop} (0.23), man (0.21), wear (0.13), \textbf{sunglass} (0.12), stand (0.09), smile (0.08), in (0.06)\\ \\\end{tabular}} & 
				
				{\tiny \begin{tabular}{@{}p{3.5cm}@{}}\\Generated captions:\\\bf A chocolate cake with white frosting on top\\ \\ Distinctive-attribute:\\ cake (0.42), chocol (0.41), plate (0.12), decor (0.12), on (0.11), \textbf{frost} (0.10), with (0.08), top (0.08)\\ \\\end{tabular}} \\
				\hline
				&
				\multicolumn{1}{c|}{\tiny (j)}& 
				\multicolumn{1}{c|}{\tiny (k)}&  
				\multicolumn{1}{c|}{\tiny (l)} \\
				\hline
				&
				\centering
				\includegraphics[width=3.5cm, height=2.2cm]{images/sm10.jpg} & 
				\includegraphics[width=3.5cm, height=2.2cm]{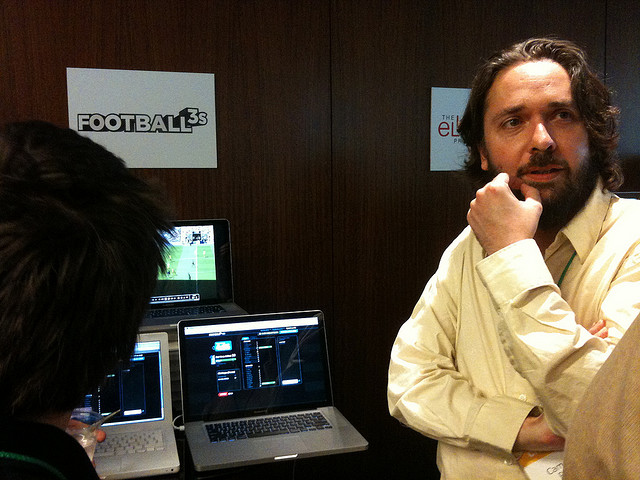} &
				\includegraphics[width=3.5cm, height=2.2cm]{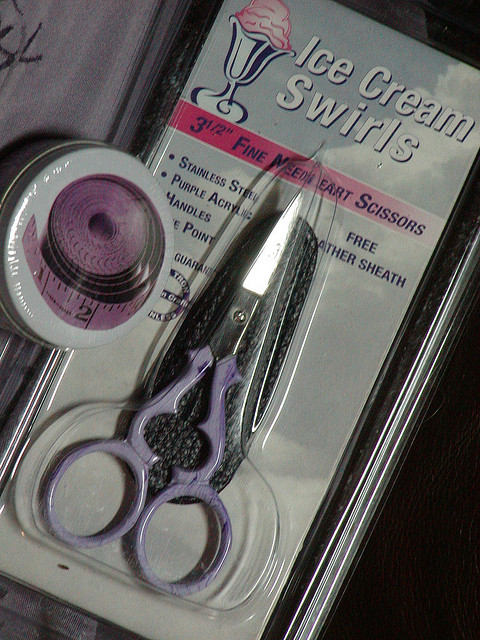} \\
				\hline
				{\tiny SCN} & 
				{\tiny \begin{tabular}{@{}p{3.5cm}@{}}\\Generated captions:\\\bf A kitchen with green walls and green walls\\ \\ Tags:\\ green (1.00), indoor (1.00), window (0.89), sitting (0.70), small (0.69), room (0.43), table (0.42), painted (0.41)\\ \\ \end{tabular}} & 
				
				{\tiny \begin{tabular}{@{}p{3.5cm}@{}}\\Generated captions:\\\bf A man holding a cell phone in his hand\\ \\ Tags:\\ person (1.00), man (0.99), indoor (0.85), front (0.66), looking (0.58), photo (0.38), standing (0.35), holding (0.31)\\ \\\end{tabular}} & 
				
				{\tiny \begin{tabular}{@{}p{3.5cm}@{}}\\Generated captions:\\\bf A cell phone sitting on top of a book\\ \\ Tags:\\ indoor (0.94), sitting (0.53), small (0.39), book (0.27), case (0.27), next (0.25), table (0.20), top (0.20)\\ \\\end{tabular}} \\
				\hline
				
				{\tiny \begin{tabular}{@{}m{0.6cm}@{}}\\DaE\\\\+\\\\SCN-\\LSTM\end{tabular}} & 
				
				{\tiny \begin{tabular}{@{}p{3.5cm}@{}}\\Generated captions:\\ \bf A kitchen with a sink and a microwave\\ \\ Distinctive-attribute:\\ \textbf{microwav} (0.44), \textbf{kitchen} (0.43), counter (0.23), and (0.11), green (0.09), with (0.09), \textbf{sink} (0.09), oven (0.09)\\ \\\end{tabular}} & 
				
				{\tiny \begin{tabular}{@{}p{3.5cm}@{}}\\Generated captions:\\\bf A man sitting in front of a computer monitor\\ \\ Distinctive-attribute:\\ \textbf{comput} (0.36), man (0.24), phone (0.17), desk (0.13), hi (0.12), at (0.12), \textbf{laptop} (0.09), sit (0.08)\\ \\\end{tabular}} & 
				
				{\tiny \begin{tabular}{@{}p{3.5cm}@{}}\\Generated captions:\\\bf A close up of a pair of scissors\\ \\ Distinctive-attribute:\\ \textbf{scissor} (0.32), \textbf{pair} (0.13), phone (0.10), of (0.10), cell (0.07), and (0.06), on (0.06), book (0.05)\\ \\\end{tabular}} \\
				\hline
				
			\end{tabularx}
		\end{center}
	\end{table}
	\setlength{\tabcolsep}{1.5pt}	
	
	\clearpage

	\bibliographystyle{splncs}
	\bibliography{egbib}
\end{document}